%% file: root.tex
\documentclass[twocolumn,letterpaper]{IEEEAerospaceCLS}  

\usepackage{booktabs}
\usepackage{graphicx}    
\usepackage{comment}

\usepackage[noadjust]{cite}
\usepackage{siunitx}
\usepackage{xcolor,colortbl}

\usepackage[figuresright]{rotating}
\usepackage{tabularx}
\usepackage{gensymb}
\usepackage{orcidlink}

\newcommand{\ignore}[1]{}  

\definecolor{green}{rgb}{0.1,0.1,0.1}
\definecolor{lightgray}{gray}{0.9}

\newcommand{\chigh}{\cellcolor[rgb]{0.7,0.9,0.7}High}
\newcommand{\cmid}{\cellcolor[rgb]{1,0.9,0.6}Mid}
\newcommand{\clow}{\cellcolor[rgb]{0.9,0.6,0.6}Low}

\begin{document}
\title{Detecting Grasping Sites in a Martian Lava Tube: Multi-Stage Perception Trade Study for ReachBot}

\author{%
Julia Di \orcidlink{0000-0001-5872-5694}\\ 
Dept. of Mech. Engineering\\
Stanford University\\
Stanford, CA 94305\\
juliadi@stanford.edu
\thanks{\footnotesize 979-8-3503-0462-6/24/$\$31.00$ \copyright2024 IEEE} 
}

\maketitle

\thispagestyle{plain}
\pagestyle{plain}

\maketitle

\thispagestyle{plain}
\pagestyle{plain}

\begin{abstract}
This paper presents a trade study analysis to design and evaluate the perception system architecture for ReachBot. ReachBot is a novel robotic concept that uses grippers at the end of deployable booms for navigation of rough terrain such as caves and lava tubes.
Previous studies on ReachBot have discussed the overall robot design, number of deployable booms, and gripper mechanism design; however, analysis of the perception and sensing system remains underdeveloped. Because ReachBot can extend and interact with terrain over long distances on the order of several meters, a robust perception and sensing strategy is crucial to identify grasping locations and enable fully autonomous operation. This trade study focuses on developing the perception trade space and realizing such perception capabilities for a physical prototype. This work includes analysis of: (1) multiple-range sensing strategies for ReachBot, (2) sensor technologies for subsurface climbing robotics, (3) criteria for sensor evaluation, (4) positions and modalities of sensors on ReachBot, and (5) map representations of grasping locations. From our analysis, we identify the overall perception strategy and sensor modality configuration for a fully-instrumented case study mission to a martian lava tube: LiDAR and radar in the body, with a 3D camera at the distal end of each boom. We also identify sensors conducive to benchtop testing and further prototype hardware development.

\end{abstract}

\tableofcontents

\section{Introduction}
\label{sec:intro}

Skylights and caves on Mars and the Moon could yield  astrobiological samples, unlock answers for geological origins, and reveal a previously undiscovered subsurface world. While there is clear scientific value in such environments, exploring them is a daunting technical challenge: navigating perilous, unknown terrain requires an innovative solution capable of versatile mobility through loose material and unpredictable obstacles. ReachBot is a robot concept that answers this call for mobility in extreme environments.

\begin{figure}
\includegraphics[width=3.25in]{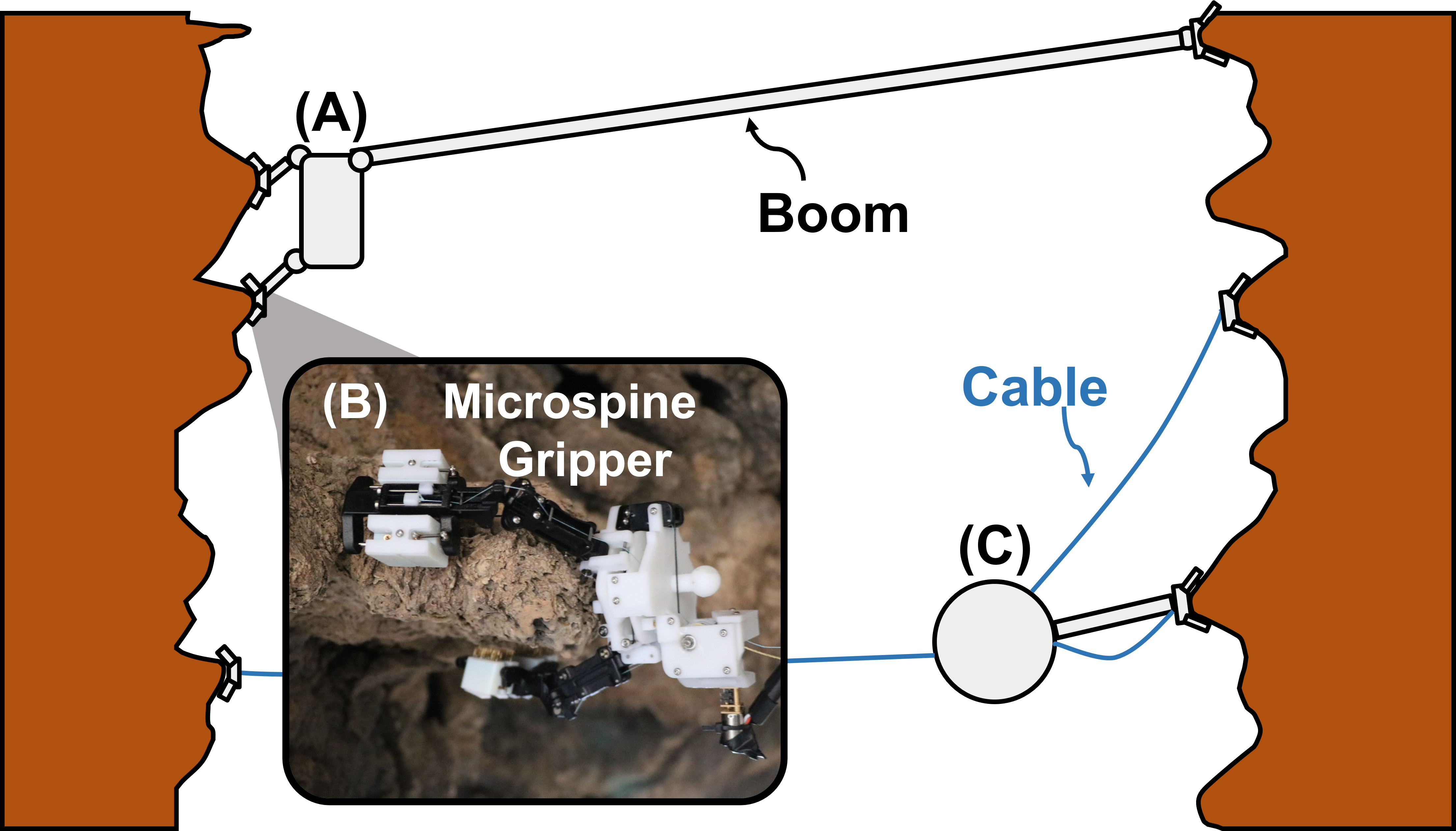}
\caption{A diagram illustrating different ReachBot configuration concepts for a martian lava tube mission. (A) shows a ReachBot configuration concept using extendable booms. An eight-boom configuration has been explored in previous work. At the end of each boom is a microspine gripper that can grasp onto rough surfaces. (B) is an image of a ReachBot microspine gripper previously tested on lava tube rocks. (C) shows a ReachBot configuration that combines cables with booms for placing the grippers.}
\label{fig:reachbotconcepts}
\end{figure}

As discussed in previous work~\cite{schneider2022reachbot,chen2022reachbot,newdick2023designing,newdick2023motion,roberts2023skeleton}, ReachBot's limbs are extendable booms that act as prismatic joints to achieve a large workspace compared to traditional rigid-link designs. These steerable booms, when paired with grippers, enable ReachBot to climb and span steep verticals inside caverns that a jumping robot or aerial robot may fail to access, as shown in Fig.~\ref{fig:reachbotconcepts}~(A,B). Furthermore, the booms could also be combined with other technologies, such as cables or tethers as shown in Fig.~\ref{fig:reachbotconcepts}~(C), to further increase reach and reduce mass and volume complexity. ReachBot's large workspace expands the number of reachable anchor points in the environment, which is especially advantageous in unknown environments with sparse anchor points or obstructions in terrain.

To grasp the environment, the gripper at the end of each boom is equipped with microspines. Microspines have been used in multiple works to grasp rough and rocky surfaces, including for planetary exploration~\cite{asbeck2006scaling,parness2017lemur,wang2019spinyhand,backus2020design,pope2016multimodal}, and several works have analyzed the conditions for successful grasping with microspines~\cite{jiang2018stochastic,wang2019spinyhand,asbeck2012designing}. As shown in Fig.~\ref{fig:reachbotconcepts}~(B), the ReachBot gripper successfully grasps surface locations that are approximately convex.

While ReachBot's mechanical architecture and motion planning have been studied before, analysis of the perception system trade space has been comparatively scant. Early work indoors assumed known grasping locations as captured by overhead cameras~\cite{chen2022reachbot}. Later, a two-stage perception strategy and sensor system was introduced in a successful field testing demonstration in a Mojave Desert lava tube~\cite{chen2024locomotion}, but this only used RGB-D images in daylight and did not review other sensing modalities or mission-level considerations. Although RGB-D images work in the field, alternatives should be investigated for more realistic mission conditions. To operate in caves and skylights, every perception system must face the typical subsurface concerns---low power, limited communication bandwidth, total darkness, and dusty conditions. These are in addition to and compounded by the operational challenges of any space hardware---radiation hardening, thermal requirements, mass/volume restrictions, cost, and technological risk. Because ReachBot has a much larger workspace than that of a typical climbing robot, examination of high-level sensing strategies is also warranted. 

\subsubsection{Statement of Contributions}

This work evaluates the perception system approaches that enable ReachBot to identify grasping sites, building upon previous ReachBot analysis as well as existing work for subsurface robots that use distance-based sensing. The main contributions of this work are:
\begin{enumerate}
    \item We present a survey of different robotic sensing modalities for a climbing robot in a subsurface environment.
    \item  We discuss perception system strategies and candidate map outputs of a ReachBot perception system.
    \item We propose evaluation criteria of sensor hardware for a configuration of ReachBot that discerns grasping locations in a martian lava tube environment.
    \item We conduct trades and propose a hardware configuration that could be physically prototyped for testing in a lava tube test environment.
\end{enumerate}

\subsubsection{Paper Organization} The rest of this paper is organized as follows. Section~\ref{sec:background} describes background literature on the perception systems of robots designed for either climbing with microspines or subsurface exploration. We also discuss considerations for upstream trades around choice of perception strategy, and downstream trades on choice of environment representation for the planner. Section~\ref{sec:tradestudy} discusses the trade study process. Then we develop evaluation criteria for sensor selection in Section~\ref{sec:componentanalysis} and present the viable perception architectures in Section~\ref{sec:sensorresults} for a specific ReachBot mission. We follow with analysis of the different map representation choices in Section~\ref{sec:graspingrepresentations}. Finally, we provide conclusions and suggest next steps for this work.

\begin{figure*}
\centering
\includegraphics[width=6in]{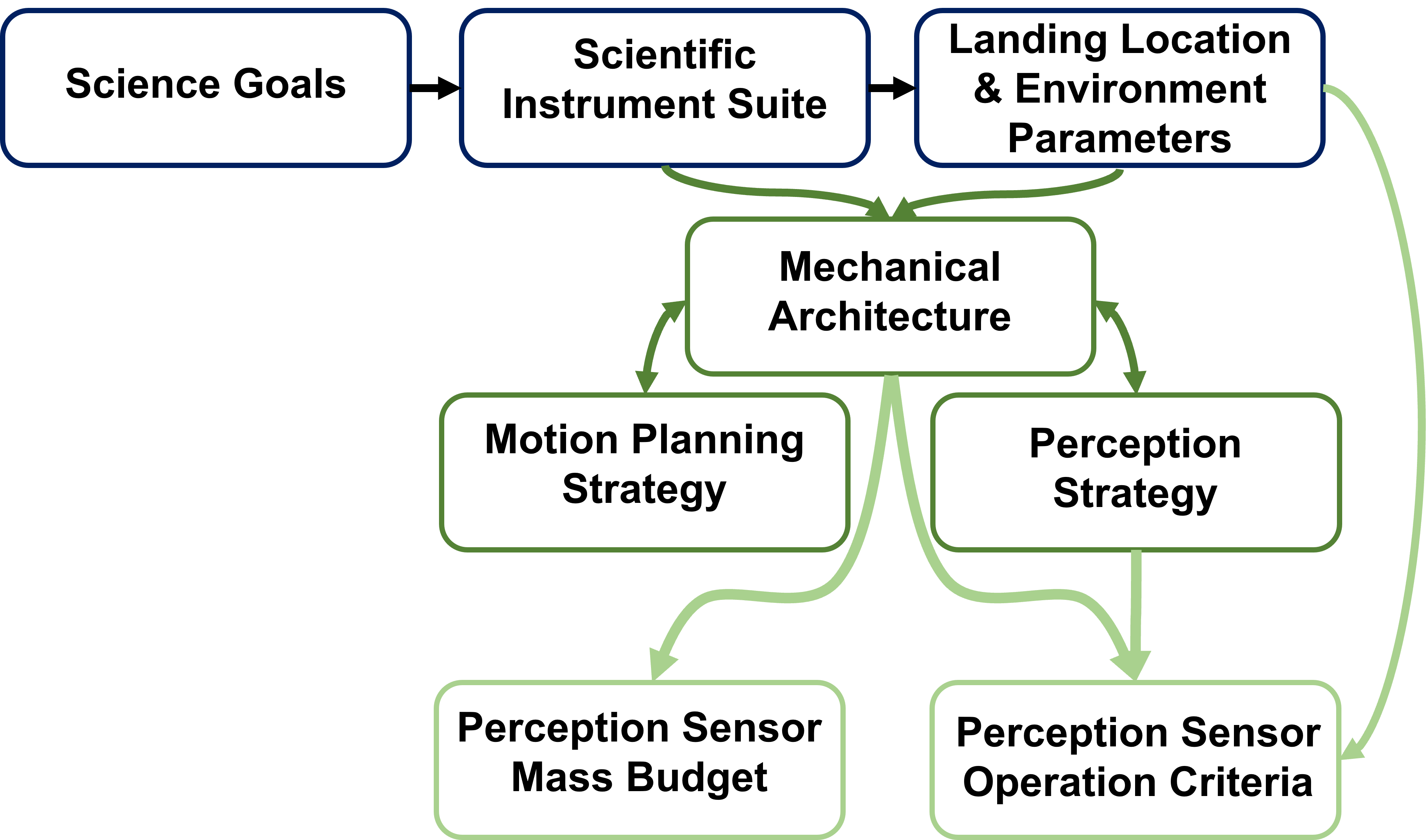}
\caption{A flowchart depicting the trade study process in this work detailed in Section~\ref{sec:tradestudy}, which begins with the dictation of Science Goals for the mission. Initial engineering trades (dark green) are derived from science parameters, with perception requirements (light green) further derived downstream.}
\label{fig:tradestudyflowchart}
\end{figure*}

\section{Background}
\label{sec:background}

\subsection{Perception Considerations for Microspine Robots}

Others have built robotic systems using microspines for climbing. Most notably, LEMUR 3 from JPL is a multi-limbed microspine-enabled robot that was field tested in a lava tube environment; their perception system was an actuated LiDAR system on the body that constructed a map of grasping locations~\cite{parness2017lemur,uckert2020investigating}. However, LEMUR and its variants are all climbing robots that hug the rock wall, thereby enabling body-mounted sensors to return highly dense and informative geometric reconstructions. ReachBot in contrast climbs by pulling on the environment with potentially extra-long limbs. A body-mounted sensor on ReachBot will almost always return much sparser data compared to a more traditional climbing robot due to the increased distance from the rock walls. Therefore, designing the perception system for ReachBot requires some attention when involving scene reconstruction from afar.

\subsection{Perception for Exploration in Subterranean Environments}

Perception systems for autonomous navigation in confined and subterranean spaces have been carefully evaluated, most recently with the DARPA Subterranean Challenge, where teams of robots searched for and detected artifacts placed in an unknown subterranean environment~\cite{chung2023into}. There are a number of relevant surveys and challenge retrospectives~\cite{tranzatto2022cerberus,agha2021nebula,ebadi2022present,hudson2021heterogeneous,azpurua2023survey} that further discuss the learnings from this Challenge.

Of import is that the majority of teams used a combination of LiDAR and IMU units for primary sensing. LiDAR units return accurate long-range depth measurements and are illumination invariant; IMUs are not sensitive to environmental disturbances and help distinguish between places with the same perceptual appearance~\cite{tranzatto2022cerberus}. Various other types of sensing (e.g. vision, thermal, odometry) are used in addition to LiDAR to increase redundancy especially when there are adverse conditions~\cite{ebadi2022present}. Multimodal perception, such as LiDAR with another modality, continues to be investigated for subsurface robotics~\cite{miki2022learning,dharmadhikari2021autonomous}. Many teams also designed modular perception payloads, allowing for standardized calibration and decoupling the system from other hardware. 

\subsection{ReachBot Perception Strategy Considerations}
 
ReachBot's motion planner relies on the perception system to generate a map of viable grasping sites. While conventional robots typically house their perception suite in the main body, ReachBot's large workspace necessitates more consideration to perception strategy. Because booms lose pointing accuracy when extended long distances, making the correction of large positional errors cumbersome, a preliminary two-stage perception strategy was proposed~\cite{chen2024locomotion}. This involved a single RGB-D camera mounted on a boom, which first perceived regions of interest from afar (boom in a stowed position), and then finalized the grasping site locations when the boom deploys closer to the surface. Here, we discuss the perception strategies for ReachBot more generally.

A one-stage process involves perception from a single vantage point on the robot, such as a perception sensor suite only mounted on the main body of ReachBot. We may assume the use of multiple sensors in this sensor suite because the body of ReachBot can hold more mass than the booms, and multiple sensors reduce the risk that ReachBot cannot perceive its surroundings. However, for ReachBots with multiple long-range booms that can extend dozens of meters in large-diameter lava tubes, the aforementioned pointing accuracy becomes an issue. Thus, the perceived spatial resolution (and accuracy) does not scale with boom length. It follows that using one-stage process is better suited for a ReachBot configuration with short booms for a narrow space such as a slim vent or crevasse, rather than a lava tube.

A two-stage perception process brings the advantage of an additional vantage point through interactive perception (perceive-act-perceive), increasing accuracy in grasping site predictions. This could take the form of a physically new sensor location on the robot or a strategically engineered one: either having a lightweight sensor mounted on the distal end of the boom to enable near-field sensing once extended, or a motion planning procedure to bring a body-mounted sensor much closer to the desired surface. Implementation of the former for multiple booms depends on the permissible mass budget of the boom and the scale of distances covered. The latter could be accomplished with more stringent stance planning, or with hybrid locomotion schemas such as rolling and climbing as part of perception.

A related question is whether there should be any gripper-level sensing, for perceiving contact. One may then implement three different range scales of sensing: far-field, near-field, and gripper-based sensing (or a subset). Instrumenting microspine grippers with force-torque, proximity, or tactile sensing could yield a number of benefits. One benefit is that estimating distance of the gripper to the surface could be used in a feedback loop to trigger the grasping sequence onto a surface (in previous tests, grasping sequences were always triggered manually). Force-torque information at the wrist could also be used for detecting gripper contact and monitoring grasping events. Sensor choice on the distal end must trade against the permissible mass budget of the boom.

\section{System Trade Study Process}
\label{sec:tradestudy}
The objective of this trade study is to determine the set of perception sensing modalities for ReachBot. We take inspiration from space engineering trade study methodologies~\cite{ziemer2018exploring} to trace architectural configuration variables from science mission objectives, a process described in Fig.~\ref{fig:tradestudyflowchart}. These variables are not exhaustive, but we use them to make initial trade decisions to narrow the space, which are summarized in Table~\ref{tab:archvariables}.

\subsection{Science Goals}
All technical specifications, including the perception strategy and sensor choices, should support a clear scientific mission. Previous papers have studied and proposed several potential top-level science missions for ReachBot~\cite{quinones2023exploring} based on the key questions highlighted in the most recent Planetary Decadal Survey~\cite{national2022origins}. We summarize the missions below:

\begin{itemize}
    \item Layered stratigraphy measurements of escarpments in the Polar Layered Deposits (PLDs) of Mars 
    \item Layered stratigraphy measurements along volcanic cliffs.
    \item Exploration of subsurface conditions in lava tubes or caves
\end{itemize}

The perception system design depends on the specific environment selected. For example, a polar region expedition will have to account for potentially highly specular icy walls, and need ground testing of sensors in these expected conditions. For this work, we focus on the lava tube and cave exploration mission, which is detailed in~\cite{quinones2023exploring} and builds upon the existing mechanical and motion planning system design analyzed in~\cite{newdick2023designing}. As a result of this mission choice, we anticipate perception system trades on operation in poor illumination and operation in dusty conditions.

\subsection{Instrument Suite}

Furthermore, the set of measurements defined in the Science Traceability Matrix from~\cite{quinones2023exploring} dictate the instrument and sensor suite. The selected scientific instruments include complementary spectrometers and a radiation detector, leading to a $25\%$ margined mass of \SI{15.1}{\kilo\gram}. We note that any additional sensors such as optical cameras could have a dual scientific and engineering use. For example, visible spectrum or Light detection and ranging (LiDAR) data used for mapping could also be used to determine geological features. Scientific instruments are located on the main body of the robot and we do not consider alternatives in this work. 

\subsection{Terrain Parameters}
Because martian lava tubes and caves have not been explored before, and because the subsurface precludes satellite pre-mapping, we assume the terrain to be uneven and unknown. Previous work has already described different terrain topologies that ReachBot concept may encounter on a mission~\cite{newdick2023designing}. We briefly list them here: 
\begin{itemize}
    \item ReachBot spanning a corridor, where the cavity is smaller than the span of the robot.
    \item ReachBot against a wall, which is mostly vertical or overhanging.
    \item ReachBot against the floor, where non-zero gravity is pushing the robot to the ground.
\end{itemize}

The perception system must perform well in all three expected terrain topology conditions. Based on previous surveys on Mars, we assign the dimensions of a representative lava tube as \SI{30}{\meter} depth, and \SI{300}{\meter} width on average~\cite{SauroPozzobonEtAl2020}. Because lava tubes and caves on Mars have not yet been explored, we do not have measurements for the expected density of anchor points. Towards building a physical ReachBot prototype, we assume the existence of some anchor points in a test environment, though the spatial frequency of these points may be irregular.

\subsection{ReachBot Configuration}
As depicted in Fig.~\ref{fig:reachbotconcepts}, one may envision several ReachBot concepts that make use of extendable booms capped with microspine grippers. We would like the following perception system analysis to be largely agnostic to the details of the specific ReachBot carrier robot. The only strict assumption we make is that all considered ReachBot concepts will have a central body and at least one extendable boom for placing microspine grippers onto the surface. This allows for both the body and the distal end of the boom as perception sensor placement possibilities. The central body may take any form, and booms may extend up to tens of meters~\cite{block2011ultralight}.

However, in order to provide a mass budget for the sensor suite, we do assume a $\approx$\SI{10}{\meter} boom extension length, building upon previous analysis of an eight-boom configuration.

\subsection{ReachBot Mass Budget}
The mass budget of the central body and distal end of the boom will dictate what sensors may be placed there (if at all). In the following calculations, we assume any wiring masses to be trivial. 

Previous work had assigned a mass budget of \SI{10}{\kilo\gram}~\cite{newdick2023designing} for an eight-boom prototype, but had assumed booms of trivial mass; and \SI{30}{\kilo\gram} had been assigned in even earlier work~\cite{schneider2022reachbot}. For a more detailed description of the mass budget, we examine existing characterization of long, lightweight deployable booms. Fourteen-meter deployable booms for space sails have previously been tested and assigned a mass density of \SI{62}{\gram} per meter~\cite{block2011ultralight}. With this construction, a \SI{10}{\meter}-long ReachBot boom would yield a \SI{0.62}{\kilogram} total mass which scales linearly with the number of booms. For example, the eight-boom ReachBot configuration yields $\approx$\SI{5}{\kilo\gram} across all booms. 

With a conservative estimate of $\approx$\SI{22.5}{\newton} pulloff force per gripper~\cite{chen2024locomotion} and given Mars gravity conditions, we expect to be capable of carrying the \SI{10}{} to \SI{30}{\kilo\gram} mass budget range assigned previously. Assuming a \SI{30}{\kilo\gram} overall budget, and taking into account the \SI{5}{\kilo\gram} for eight booms and a margined mass of \SI{15.1}{\kilo\gram} for the required scientific sensor suite, we therefore assign $20\%$ of the remainder or $\approx$\SI{2}{\kilo\gram} for the perception sensors located on the body. We note that the eight-boom design is likely the upper threshold on number of booms (designed to achieve force closure), and that there are other configurations with fewer booms when the force closure restraint is relaxed (and therefore reducing mass). 

A mass budget for the distal end of the boom can also be derived. Specifically, the limiting case for boom failure is buckling~\cite{newdick2023designing}. The most buckling load is experienced when the boom is completely outstretched and perpendicular to the direction of gravity. In this configuration, the boom supports the weight of the boom, the weight of the gripper, and the weight of any sensor(s) at the distal end as given by: 

\begin{equation}
M_{\text{shoulder}} = (m_{\text{sensor}} + m_{\text{gripper}} + \frac{1}{2}m_{\text{boom}}) g_{\text{Mars}}L 
\end{equation}

where $M$ is the buckling moment felt at the shoulder, $m$ is mass, $g_{\text{Mars}} = \SI{3.71}{\meter\second^2}$, and $L = 10$ meters. A critical buckling moment of $M = \SI{59.8}{\newton\meter}$ was reported in for the 14-meter deployable space sail boom~\cite{sickinger2006structural,block2011ultralight}. Applying a $25\%$ margin on the buckling moment, and using $m_{\text{boom}} = \SI{0.62}{\kilogram}$ and $m_{\text{gripper}} = \SI{0.25}{\kilo\gram}$ (gripper mass reported in~\cite{chen2024locomotion}), we find that the sensor(s) may have a mass up to \SI{0.72}{\kilo\gram} when located at the distal end of the boom.

\subsection{ReachBot Perception Strategy}

As analysed in Section~\ref{sec:background}, a two-stage perception process has the benefit of additional surveying, and is preferred over a one-stage perception process especially when ReachBot is in an environment with potentially wide expanses to cover. Based on the mass budget analysis, we are able to accommodate sensing at the body and at the distal end of the boom, thus allowing for a two-stage interactive perception strategy. We leave open the possibility of force sensing at the gripper and defer the specifics to future work. 

We define the "near-field" distance as one-third of the length of the boom; for example, for a 10-meter boom, the range for near-field sensing is \SI{3.3}{\meter} or closer to the surface.

\begin{table*}[t]
  \centering
  \rowcolors{2}{}{lightgray}
    \caption{We make architectural trades to narrow the trade space, as summarized in this table.}
    \label{tab:archvariables}
  \begin{tabularx}{\textwidth}{lX}
  
  \cellcolor[rgb]{1,0.8,0.2}{\textbf{Trade Space Variable}} & \cellcolor[rgb]{1,0.8,0.2}{\textbf{Selected Definition}} \\
  \hline
    Environment Location & Lava tube or cave within the Tharsis Montes (Arsia Mon)  region of Mars~\cite{cushing2007themis} \\
    
    Terrain Parameters & Lava tube or cave up to \SI{30}{\meter} depth and \SI{300}{\meter} width~\cite{SauroPozzobonEtAl2020}. \\
    
    Instrument Suite Location & Science instruments are only on the ReachBot body. \\
    
    ReachBot Configuration & Concept includes one central body and at least one extending boom that may place microspine grippers onto a grasping location. Booms extend \SI{10}{\meter}. Wiring mass trivial. \\
    
    ReachBot Mass Budget &
    Up to \SI{2}{\kilo\gram} body-based sensor(s), \SI{0.72}{\kilo\gram} gripper-based sensor(s) with $25\%$ margin. Assume any power cables and wires have trivial mass. \\
    
    ReachBot Perception Strategy & Two-stage with possible gripper-level force sensing. 
    Near-field is defined as the distance of one-third of the boom length. \\
    \hline
  \end{tabularx}

\end{table*}

\section{Perception Sensor Selection}
\label{sec:componentanalysis}
 To avoid the time and cost of sensor development, this perception system trade study is interested in analyzing only commercial-off-the-shelf (COTS) or mature laboratory technology for a prototype. Because we aim to prototype a full ReachBot system eventually, we also do not consider highly expensive sensors in this survey. In some cases, modifications to sensors are proposed to increase performance or range for comparison. In the following, we summarize common range-sensing technologies and provide a list of criteria for evaluation for a ReachBot perception system. As part of the trade study process, we also investigate the benefits and drawbacks of different sensor modalities.

\subsection{Range-Sensing Technologies for Subsurface}
Unlike wheeled or quadruped robots, which interact mostly with the ground floor, ReachBot and other such climbing robots will have significantly more interaction with the walls and ceilings of an environment. Several works have studied and compared perception sensor categories for subterranean robotics~\cite{wong2011comparative,leingartner2016evaluation,azpurua2023survey}, though to our knowledge this is the first such analysis for a subsurface robot with a multiple-stage perception strategy. In this work, we consider the following range-sensing technologies: LiDAR, 2D cameras (monocular), 3D cameras (RGB-D or Time of Flight (ToF)), sonar, radar, and thermal. 

\subsubsection{LiDAR}
LiDAR sensors are commonly used in robotics for mapping the environment. LiDAR sensors measure ToF of short laser pulses. These pulses can be rastered using spinning mirrors to generate a point cloud of the surroundings. More specifically, the ToF data is dependent on and thus relates to the angle, distance, and reflectivity of the surface. We note that there are other types of LiDAR technologies such as focal plane array (Flash) LiDAR, frequency modulated continuous wave (FMCW) LiDAR. These have very long range (beyond \SI{150}{\meter}), but are generally low spatial resolution at a distance. The low data density can be mitigated by mounting the LiDAR unit on a motorized tilt platform for sweeping samples. Another drawback of LiDAR is degraded performance in dusty or foggy conditions which may occur in the subsurface. However the major benefit is LiDAR does not need external illumination. Because we plan to operate in a 3D terrain environment, we are only interested in 3D LiDAR.

\subsubsection{2D Cameras}
Monocular cameras represent the scene as a 2D array of pixels, which can either be color or grayscale. It is possible to detect depth using structure from motion or other techniques, and planar images have been used in combination with LiDAR to estimate motion. For example, high-frame-rate event cameras report can changes at the microsecond scale~\cite{gallego2020event}. Monocular cameras also will require onboard illumination to function in a subterranean environment, which could introduce issues with shadows, brightness differences, specular reflections, and occlusions. This could make visual and visual-inertial mapping difficult downstream.

\subsubsection{3D Cameras}
Depth cameras such as RGB-D or stereo can produce 3D or depth data as well as traditional images. Stereo cameras work by extracting depth information from multiple monocular cameras that are arranged with known extrinsic/intrinsic parameters. RGB-D cameras, such as the Intel Realsense, provide both color and depth information. Active RGB-D cameras project patterns of structured infrared light into a scene and use the sensed deformation of the patterns of objects in the scene to accurately estimate depth, which is more robust to lack of proper illumination. Despite this, the IR light and thus depth measurements have degraded performance in brighter outdoor environments. More discussion on active 3D vision cameras can be found in~\cite{horaud2016overview}.

\subsubsection{Sonar}
Sound navigation and ranging (SoNAR) sensors measure ToF of ultrasound waves. SoNAR trades off short-range precision with spatial accuracy, and has been found to fare better in humidity than LiDAR~\cite{silver2004scan}. Like LiDAR, SoNAR does not require active illumination. There exists several commercially-available sonar sensors.

\subsubsection{Radar}

Radio detection and ranging (RADAR) is another active range sensing technology. As with SoNAR, RADAR does not require light or temperature gradients to operate, and system-on-chip (SoC) radars have low power draws. However, radars can be adversely affected by sensor noise, spatial resolution, and data corruption (e.g. multipath reflections)~\cite{kramer2020radar}. Millimeter-scale radar waves are also large enough to be less affected by small airborne particulates, which can cause reflections with LiDAR; however, radar is not as common as camera vision on subsurface robots today~\cite{ebadi2022present}.

\subsubsection{Thermal}
Thermal vision detects the magnitude of infrared radiation which can be output as a 2D heat signature image. Thermal sensing has previously been shown for ground granularity prediction in wheeled robot navigation~\cite{cunningham2019improving}. However, the usefulness of thermal vision depends on differences in temperature; while on the surface, thermal transients are expected due to periodic illumination, in subsurface caves such gradients are often minimal. It is noted that thermal may be useful if there is water present, or if there is an opening in the ground, especially as thermal vision is not affected by dust degradation~\cite{dharmadhikari2021autonomous}.

\input{tabmodalities}

\subsection{Sensor Evaluation Criteria}

Trade studies require criteria to evaluate both effectiveness and performance, so we list the following criteria: 

\begin{enumerate}
    \item \textbf{Spatial Resolution}. As a requirement, we need enough spatial resolution to distinguish grasping site features. The smallest graspable surface would involve a pinch grasp, and we model this surface as a \SI{50}{\milli\meter} diameter hemisphere as based on the ReachBot palm size \cite{chen2024locomotion}. Therefore we require a spatial resolution of \SI{25}{\milli\meter^2} per measurement or better when near-field sensing (this can be obtained through multiple sensors or sensor fusion techniques if not outright). If the minimum resolution at the near-field sensing distance is achieved, the objective is to have high spatial resolution at a distance; the higher the spatial resolution the better the score. We may also weigh this score by sensor range to optimize for sensors with high resolution at mid to long ranges.

    \item \textbf{Spatial Accuracy}. Accurate distance measurements are vital for precise navigation and manipulation. For this trade study we do not list a specific accuracy requirement but have this criteria as an objective: the more accurate the sensor, the higher the score.

    \item \textbf{Range}. The detection range impacts ReachBot's ability to detect distant obstacles or grasping sites, so range capabilities are required for the sensors. For a near-field sensor, we have defined near-field previously as being one-third of the total boom length, so we require sensing at least from \SI{3.3}{\meter} for a \SI{10}{\meter}-long boom. We assign a higher score if able to sense at a closer distance. For a far-field sensor, we require a minimum detection range from the one-third length to the full length of the boom. Given the lava tube environment dimensions (Table~\ref{tab:archvariables}), we assign a higher score for more distance, though with no need beyond \SI{20}{\meter}. To avoid blind spots in sensing range, there should be some switchover bandwidth built in when switching from far- to near-field sensing.

    \item \textbf{Field of View}. This is an objective to maximize field of view; there is no minimum field of view required. Because ReachBot must be able to map not only floors, but walls and ceilings, we assign a higher score for sensors with higher field of view. We note that some sensors may be mounted on mechanisms to increase their field of view, such as a motorized platform.

    \item \textbf{Robustness to Darkness}. Sensors must withstand harsh planetary environments to ensure long-term reliability and minimize maintenance needs. Sensors are required to be able (or can be augmented) to operate in consistent total darkness. 

    \item \textbf{Robustness to Dust}. Sensors are required to be able (or can be augmented) to operate in  dusty conditions.

    \item \textbf{Power Efficiency}. Energy-efficient sensors help maximize mission duration and reduce the need for frequent recharging. Power efficiency is an objective not a requirement, so the more power efficient, the higher the score.

    \item \textbf{Implementation Ease}. The ideal sensor would levy no constraints on the design of the rest of the system, are supported by the manufacturer and the scientific robotics community, and require no additional modifications to operate. Higher scores are given to sensors with recent and widespread heritage in subsurface robotics and are still supported.

    \item \textbf{Lightness and Compactness}. As with any space application, minimization of mass and volume is essential. Higher scores are given to lightweight, compact sensors (objective) that fit within the ReachBot mass budget (requirement).

    \item \textbf{Affordability}. This metric was included to favor options that could be inexpensively prototyped and tested. The more inexpensive and easy to obtain, the higher the score.
\end{enumerate}

We note that there are many other evaluation criteria that could have been chosen, such as robustness to extreme temperatures, radiation resistance, or computation efficiency. The ultimate goal for this trade study is to use this analysis to inform a feasible perception system prototype; if ReachBot is selected for a space mission then a more exhaustive study can be done in the future to include space hardiness parameters. As a result, we assign a weight of 2 to performance-based criteria (Resolution, Accuracy, Range, Field of View, Robustness to Darkness, Robustness to Dust) and a weight of 1 to the space-worthiness criteria (Power Efficiency, Implementation Ease, Lightness and Compactness, Affordability).

\section{Sensor Trade Results}
\label{sec:sensorresults}

Given multiple range-sensing technologies and ReachBot-specific evaluation criteria, we now present the results of our trade study. 

\subsection{Sensor Modality Trades for ReachBot}

\input{tabsensors}

To fully instrument a ReachBot for a lava tube mission, we conduct a trade to select the primary modalities for ReachBot and report these results in Table~\ref{tab:modalities}. More detail on the construction of this table is reported in the appendices.

First, we analyse Table~\ref{tab:modalities} for the best modality in far-field sensing (scored "High" in range) and find LiDAR to be the best sensor type across our desired characteristics for far-field sensing. This aligns with other subsurface robotics literature, which generally also have LiDAR-centric perception systems~\cite{ebadi2022present}. Because of its capabilities over long ranges, we choose LiDAR as the primary far-field sensing solution.  

Since ReachBot must be able to see both the floor and ceiling, the LiDAR unit will be mounted on the body of ReachBot and may have additional supporting systems. For example, one supporting system pairs the LiDAR with an IMU on a spinning, tilted mount, similar to the CatPack pioneered by CSIRO's Wildcat system~\cite{hudson2021heterogeneous}. This mount increases the field of view dramatically; a Velodyne VLP-16 LiDAR mounted on such a system provides \ang{120} vertical field of view instead of the usual \ang{30}, therefore allowing ReachBot to perceive the floor, walls, and ceiling of the lava tube at great distances though with added software and mechanical complexity. As another example, we may employ two LiDAR units, with one tilted for primarily ground-to-wall coverage in the forward direction, and the other for primarily wall-to-ceiling coverage in the forward direction. 

For further redundancy and resilience to edge cases, the primary LiDAR modality can be complemented with other sensing. Based on the results of Table~\ref{tab:modalities}, we consider also pairing LiDAR with radar, which is able to cover long ranges and is also robust to dusty conditions in cases where LiDAR may fail. An example radar sensor is the Texas Instruments AWR1843 sensor, which has been previously used for robotics~\cite{kramer2022coloradar,kramer2020radar}. However, we note that the integration of radar sensing is not as established as other sensing modalities because of the complexity of the corresponding sensor models and data, so more research must be done to tightly couple the two modalities. 

For the near-field sensing (ranked "Low" or "Mid" in range), 3D cameras perform best. This sensor will be mounted at the distal end of the boom. An example of a 3D camera is the Intel Realsense D435i, which has seen widespread use in robotics. One additional benefit to having a RGB vision capability at the distal end of the boom is the ability to capture color images of geological features for scientific study. This will require onboard active illumination at the sensor itself for color images, and further analysis can be done on the illumination design.

Therefore, from a modality standpoint, a fully-instrumented ReachBot with a multimodal perception suite should consist of LiDAR and radar located in the body and a 3D camera located at the distal end of each boom. One possible ReachBot body system that fulfills these modalities would be two Velodyne VLP-16 pucks mounted at opposing \ang{45} angles and forward- and rear-facing Texas Instrument AWR1843 System on Chip automotive-grade radar sensors (all together $\approx$\SI{1.7}{\kilo\gram}).

\subsection{Sensor Component Trade for a Prototype}

In the end, we would like to build a physical prototype of the perception system. Towards this, in Table~\ref{tab:sensors}, we report the specific evaluations of several commercially-available sensors\footnote{\bf{We note that in the case of the Apple iPhone 12 LiDAR, detailed technical information is not publicly available, so we use approximate values reported in literature~\cite{luetzenburg2021evaluation}.}} for ReachBot based on the evaluation criteria. These sensors were chosen because they are either already in our possession, or are used previously in literature and purchaseable online. The full treatment with all parameters and details is included in the appendices.

For LiDAR-based far-field sensing, we currently possess a Velodyne VLP-16 unit, but would like to compare this unit against alternative LiDAR units that are purchase-able online, in case they perform better. Since we already have a Velodyne unit, in the trade we give "Affordability" full marks; because LiDAR units are quite costly we assign a weight of "2" for the "Affordability" metric. Since we want to be able to quickly prototype, we also assign a weight of "2" for the "Implementation Ease" metric. The results of this comparison is shown in the far-field sensing section of Table~\ref{tab:sensors}.

Based on the weighted scoring, the Velodyne VLP-16 and Ouster OS1-32 unit are both good options for ReachBot's primary LiDAR unit. The Ouster OS1-32 outperforms the Velodyne unit on both resolution and field of view, but costs over \$6000 to purchase online, which is quite prohibitive. Unless a unit is available for a cheap price, we will develop using the existing Velodyne VLP-16 unit in our possession.

For vision-based near-field sensing, we currently possess an Intel D455 RGB-D camera that was used in field tests~\cite{chen2024locomotion}, and an Intel D435i. We would like to compare them against alternative RGB-D cameras and investigate which option fares best given a better understanding of the sensing evaluation requirements for near-field sensing. The results are shown in near-field sensing section of Table~\ref{tab:sensors}.

Based on the weighted scoring, the Intel D435i and the StereoLabs Zed2 are both good options for ReachBot's primary near-field sensing unit. The Intel D435i, as a commonplace sensor in robotics with many open-source libraries for realsense software, scored highest for implementation ease. Based on this trade study, we recommend using the Intel D435i for near-field sensing. However, the Zed2 has a wider field of view, a wider range, and is \SI{100}{\gram} lighter than the Intel D435i, and presents a good second option.

The result of this trade study is a recommendation for the specific primary component for both far- and near-field sensing. Based on this analysis, we will use a Velodyne VLP-16 unit for far-field sensing and an Intel D435i for near-field sensing. We have also identified promising alternatives to both components should product cancellations or other external vendor delays occur.

\section{Grasping Site Representation Analysis}
\label{sec:graspingrepresentations}

\begin{figure}
\includegraphics[width=3.25in]{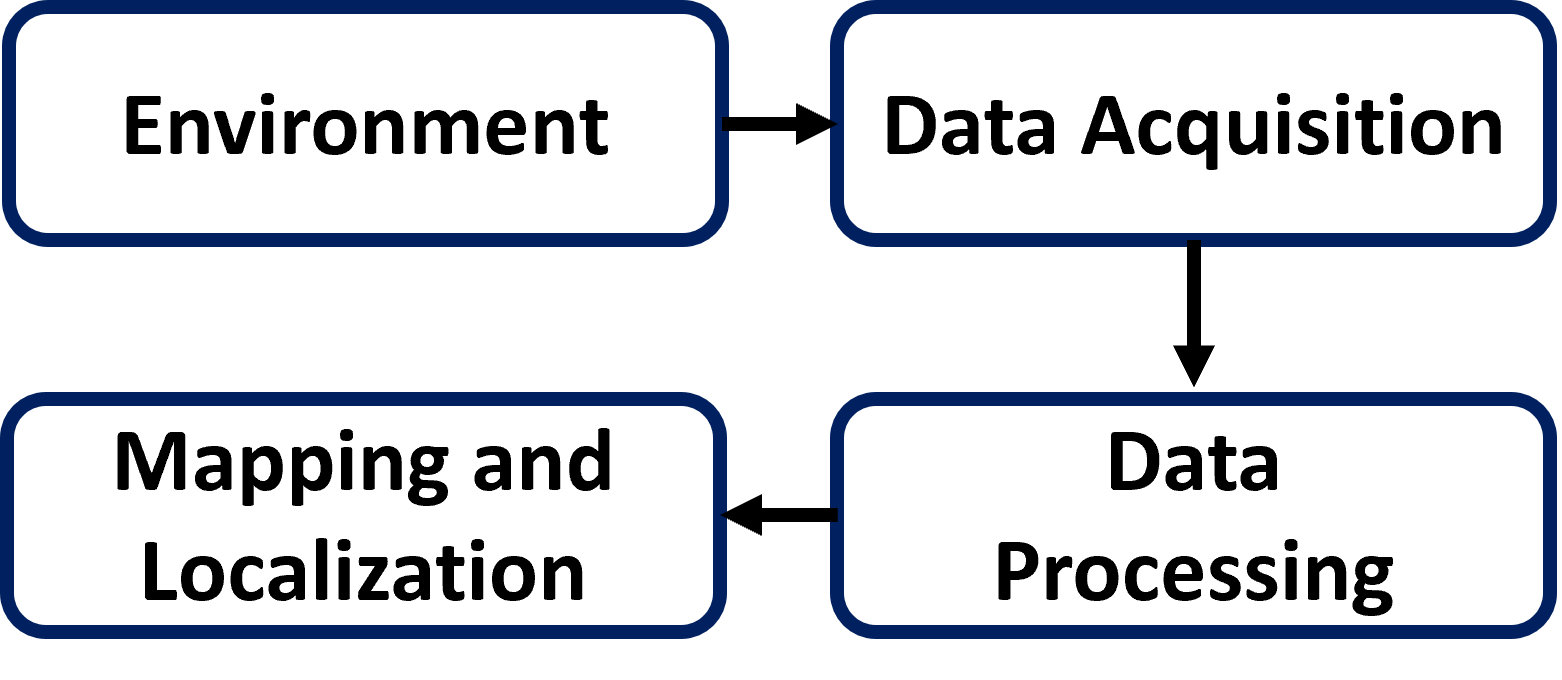}
\caption{A flowchart of perception system for ReachBot: with the environment as input, the system acquires data, processes data, and outputs mapping and localization products for the planner. The majority of this trade study focused on defining hardware constraints and optimization objectives (hardware, sensors). We also discuss different representation possibilities for ReachBot downstream in order to provide a full picture of the perception system.}
\label{fig:perceptionflow}
\end{figure}

As shown in Fig.~\ref{fig:perceptionflow}, the goal of the perception system is to provide a map output (of the grasping sites) for the planner. We have thus far developed the primary and secondary modalities for ReachBot, and suggested specific sensor components for further prototyping. Now we will provide some analysis of the software architecture for ReachBot's perception system, but leave specific recommendations to future work.

At a high level, the choice of map representation should provide compatible inputs for the planner. For identifying grasping site locations, a force-torque limit surface can be generated given known gripper parameters and macro rock geometry with conservative rock friction properties assumed. Macro rock geometry is perceived from the sensors discussed in the previous section, and can be parameterized (curvature, radius, centroid of location) and stored in the terrain map. The computation of a limit surface is somewhat time consuming (highly dependent on contact details of microspines on rock surfaces), but could be pre-computed offline for typical cases or learned from representative cases. This would likely be implemented as an additional layer when processing perceptual data to output to the planner.

Terrain representation types can generally be categorized as 2D, 2.5D, and 3D maps, with existing surveys detailing more specific options for robotic exploration~\cite{azpurua2023survey}. The choice of map type trades accuracy and resolution with computational efficiency and memory storage. Because a lava tube is a highly unstructured outdoor environment, we focus only on analyzing 2.5D and 3D map representations here.

\subsection{2.5D Maps}
The most common 2.5D mapping method is an elevation map, where $(x,y)$ locations are augmented with a height $h$ of the surface. Naturally, elevation maps make sense for environments where heights can be computed, such as the floor of a mine shaft or any such space with a guaranteed ground floor. The ANYmal platform uses elevation maps for real-time surface reconstruction, called GridMap~\cite{fankhauser2018anymal}. 

\subsection{3D mapping}

Pointclouds represent occupied points but are generally not used for real-time planning. OctoMap is a more efficient representation of occupancy probabilities, but using Octrees (a tree data structure) to divide a 3D space into octants~\cite{meagher1982geometric}. OctoMaps allow for empty and unexplored spaces to be represented. Another representation is to use Truncated Signed Distance Functions (TSDF), which is a technique from computer graphics that measures the distance between each point in a point cloud and the surface of an object being represented~\cite{oleynikova2017voxblox}. Another representation can be a mesh which defines an object shape as a series of triangular meshes, which have highly descriptive capabilities at expense of computation cost~\cite{himmelsbach2010fast}.

\subsection{Discussion and Suggested Criteria}
As with any real robot system, we would like to keep memory and computation costs low. There are also some criteria that are specific for ReachBot's tasks: being able to represent multiple scales of resolution, and being able to represent unstructured 3D environments. Although we leave this for a future trade study, we suggest a few usability criteria for map representations in confined spaces:

\begin{itemize}
  \item Low memory usage
  \item Low computation overhead
  \item Multiple-scale environments (especially with respect to near- and far-field sensing)
  \item Multiple resolutions (especially with respect to near- and far-field sensing)
\end{itemize}

\section{Conclusions}
\label{sec:conclusions}

In this work, we analyze perception system design tradeoffs and develop the configuration of sensor modalities for a ReachBot mission to a martian lava tube. We also discussed overall perception strategy and different mapping representations for ReachBot. We find that a LiDAR-centric approach is the best far-field modality for a ReachBot mission; this also aligns with existing literature on LiDAR-based solutions, which have become increasingly robust to challenging environments and are common primary sensing modalities for subsurface robots. We also recommend a specific LiDAR sensor for prototyping, and find that supporting LiDAR with another complementary modality such as radar will make the system more robust to dusty conditions. Finally, we find that for near-field sensing, 3D vision such as RGB-D cameras perform best and we recommend a specific camera sensor for further development work.

There are a few notable limitations to this trade study. First, we did not provide a link budget or a discussion of data compression methods for the perception suite. This is an area that we will further investigate as part of realizing physical prototypes of this system. In addition, although we give an estimate of the mass budget, we do not give an estimate of the power budget. For testing on Earth we assume access to a reliable power source, but for a true Mars mission, much more power budget analysis and optimization is necessary when designing the perception suite. This may also include investigation into different types of power sources, such as nuclear and solar, and sensor payload trades based on power needs. Finally, we also do not discuss the scientific instrument payload here. Although there is some overlap that allows engineering cameras to collect data for scientific use, we largely assume that the scientific payload (e.g., spectroscopy instruments) are separate in scope from engineering. 

Having identified leading COTS component candidates for the ReachBot perception system based on the trade study criteria, our next steps involve refining the implementation details. These include hardware testing and characterization, investigation into illumination approaches to support the near-field vision sensing, and further development of the grasp site representation and reconstruction method in pursuit of a physical ReachBot prototype. Similar trade studies may be conducted for designing ReachBot perception systems for even more adverse conditions, such as the specular ice walls of the Martian poles or Europa!

\appendices{}              


\section{Modality Analysis}
\label{app:modality}

We prioritize performance-based criteria (Resolution, Accuracy, Field of View, Range, Robustness to Darkness, Robustness to Dust) by a weight of 2, and others by a weight of 1. 

Here are the cut-off ranges selected for the descriptions in Table~\ref{tab:modalities}.
\begin{itemize}
    \item \textbf{Resolution}: We consider high resolution to be capable of returning 2MP or more. We define low resolution as only able to return 0.5MP or less.
    \item \textbf{Accuracy}: We consider high accuracy to be less than or equal to 2\% error at nominal range, and low accuracy to be greater than 10\% error at nominal range (as reported by the manufacturer).
    \item \textbf{Field of View}: We consider high field of view to be greater than \ang{90} in either vertical or horizontal field of view without any additional support (i.e. special motorized mounts). 
    \item \textbf{Range}: We consider greater than the nominal boom length of \SI{10}{\meter} as high range. We consider less than \SI{3}{\meter} as low range.
    \item \textbf{Power Efficiency}: We consider high power efficiency to be \SI{}{\milli\watt}-level power draw, mid power efficiency to be on the order of \SI{1}{\watt} power draw, and low power efficiency to be on the order of \SI{10}{\watt} power draw.
    \item \textbf{Robustness to Darkness}: We consider high robustness to darkness if the sensor is invariant to illumination. We note that in the case of 3D cameras, the RGB portion of the camera is not illumination invariant, but the depth portion is. We consider low robustness to darkness if the sensor cannot produce useable data in darkness conditions.
    \item \textbf{Robustness to Dust}: We consider high robustness if the sensor is invariant to dust. We consider low robustness if dusty conditions causes substantial decrease in data quality.
    \item \textbf{Implementation Ease}: We consider high implementation ease if the technology has been used in many subsurface robots (such as those in the DARPA Subterranean Challenge) and existing literature. We consider low implementation ease if the technology has only been used in one or two subsurface robots or requires more research.
    \item \textbf{Lightness and Compactness}: We consider "High" to be under 10 grams (implementable on a chip); and "Low" to be over 500 grams (must be mounted).
    \item \textbf{Affordability}: We consider prices in the order of magnitude of 10 USD to be highly affordability, 100 USD to be mid affordability, and 1000 USD to be low affordability.
\end{itemize}

\section{Sensor Analysis}
\label{app:fullSensor}

In the following table, Table~\ref{tab:full}, we include the full technical details for the sensors considered in this trade study. 
\input{tabfull}

\acknowledgments
Support for this work was provided by NASA under the Innovative Advanced Concepts program (NIAC) and by the Air Force under an STTR award with Altius Space Machines. Many thanks to Stephanie Newdick and Tony Chen for providing feedback on this manuscript.

\bibliographystyle{IEEEtran}
\bibliography{reachbot}

\thebiography
\begin{biographywithpic}{Julia Di}{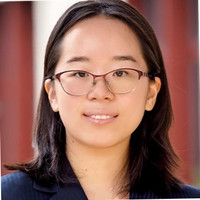}
is a Ph.D. candidate in the Biomimetic and Dexterous Manipulation Laboratory and was a NASA Space Technology Graduate Research Fellow with the NASA Ames Research Center Intelligent Robotics Group and NASA Jet Propulsion Laboratory, She received a B.S. degree in electrical engineering from Columbia University in 2018. Her research interests are vision-based tactile sensing and perception for robotic systems. Her goal in life is to explore the intersections of art and social good with technology.
\end{biographywithpic} 

\end{document}

%% file: tabmodalities.tex
\begin{table*}
\centering
    \caption{Modality Evaluation: The capabilities and basic sensor information of various sensor categories evaluated in terms of ReachBot criteria. Criteria cut-offs are discussed in the appendices.}
    \label{tab:modalities}
\begin{tabularx}{\textwidth}{l l  *{10}{X} }

\bfseries Modality & 
\bfseries Example Device &
\bfseries \rotatebox{60} {Resolution}  & 
\bfseries \rotatebox{60} {Accuracy}  & 
\bfseries \rotatebox{60} {Field of View}  & 
\bfseries \rotatebox{60} {Range}  & 
\bfseries \rotatebox{60} {Power Efficiency}  & 
\bfseries \rotatebox{60} {Robustness to Darkness}  & 
\bfseries \rotatebox{60} {Robustness to Dust}  & 
\bfseries \rotatebox{60} {Implementation Ease}  & 
\bfseries \rotatebox{60} {Lightness and Compactness}  & 
\bfseries \rotatebox{60} {Affordability}  \\
\hline
\cellcolor{lightgray}LiDAR & \cellcolor{lightgray}Velodyne VLP-16  & \chigh & \chigh & \chigh & \chigh & \clow & \chigh & \clow & \chigh & \clow & \clow \\

2D Camera & FLIR Firefly S & \chigh & \chigh & \chigh & \clow & \cmid & \clow & \clow & \chigh & \cmid & \chigh \\

\cellcolor{lightgray}3D Camera & \cellcolor{lightgray}IntelD435i & \chigh & \chigh & \chigh & \clow & \cmid & \chigh & \clow & \chigh & \cmid & \cmid \\

Radar & XM132 &\cmid & \cmid & \cmid & \chigh & \chigh & \chigh & \chigh & \cmid & \chigh & \cmid \\

\cellcolor{lightgray}Sonar & \cellcolor{lightgray}MaxBotix MB1000 &\cmid & \cmid & \cmid & \cmid & \chigh & \chigh & \chigh & \cmid & \chigh & \cmid \\

Thermal & FLIR Tau2 & \cmid & \cmid & \cmid & \cmid & \cmid & \chigh & \cmid & \clow & \cmid & \cmid \\
\hline

\end{tabularx}
\end{table*}

%% file: tabsensors.tex
\begin{table*}[t]
\rowcolors{4}{}{lightgray}
\centering
    \caption{Sensors considered for building a ReachBot prototype, where each device is scored from 0 to 2 per criterion. The scores are multiplied by the respective weight to produce the weighted sum in the rightmost column.}
    \label{tab:sensors}
\begin{tabularx}{\textwidth}{ l l *{10}X r}

\bfseries Sensor Type &
\bfseries Device Name & 
\bfseries \rotatebox{60}{Resolution} & 
\bfseries \rotatebox{60}{Accuracy} & 
\bfseries \rotatebox{60}{Field of View} & 
\bfseries \rotatebox{60}{Range} & 
\bfseries \rotatebox{60}{Robustness to Darkness} & 
\bfseries \rotatebox{60}{Robustness to Dust} & 
\bfseries \rotatebox{60}{Power Efficiency} & 
\bfseries \rotatebox{60}{Implementation Ease} & 
\bfseries \rotatebox{60}{Lightness and Compactness} & 
\bfseries \rotatebox{60}{Affordability} & 
\bfseries Weighted Sum \\
\hline

\multicolumn{13}{c}{\cellcolor[rgb]{1,0.8,0.2}Far-field Sensing}\\
\hline
\multicolumn{2}{c}{\cellcolor[rgb]{1,0.8,0.2}Scoring Weights} & \cellcolor[rgb]{1,0.8,0.2} 2 & \cellcolor[rgb]{1,0.8,0.2} 2 & \cellcolor[rgb]{1,0.8,0.2} 2 & \cellcolor[rgb]{1,0.8,0.2} 2 & \cellcolor[rgb]{1,0.8,0.2} 2 & \cellcolor[rgb]{1,0.8,0.2} 2 & \cellcolor[rgb]{1,0.8,0.2} 1 & \cellcolor[rgb]{1,0.8,0.2} 1 & \cellcolor[rgb]{1,0.8,0.2} 1 & \cellcolor[rgb]{1,0.8,0.2} 2 & \cellcolor[rgb]{1,0.8,0.2} \\
 \hline

LiDAR & RSBPearl & 2 & 2 & 2 & 1 & 2 & 1 & 1 & 1 & 1 & 0 & \textbf{23} \\

LiDAR & Velodyne Puck (VLP-16) & 1 & 2 & 1 & 2 & 2 & 1 & 1 & 2 & 1 & 2 & \textbf{26} \\

LiDAR (ToF) & Cygbot Mini Lidar & 1 & 2 & 0 & 0 & 2 & 1 & 2 & 0 & 2 & 2 & \textbf{20} \\

LiDAR & iPhone 12 & 2 & 2 & 1 & 0 & 2 & 1 & 2 & 1 & 2 & 1 & \textbf{23}  
\\

LiDAR & Ouster OS1-32 & 2 & 2 & 2 & 2 & 2 & 1 & 1 & 1 & 2 & 0 & \textbf{26} \\


\multicolumn{13}{c}{\cellcolor[rgb]{1,0.8,0.2}Near-field Sensing} \\
\hline
\multicolumn{2}{c}{\cellcolor[rgb]{1,0.8,0.2}Scoring Weights} & \cellcolor[rgb]{1,0.8,0.2} 2 & \cellcolor[rgb]{1,0.8,0.2} 2 & \cellcolor[rgb]{1,0.8,0.2} 2 & \cellcolor[rgb]{1,0.8,0.2} 1 & \cellcolor[rgb]{1,0.8,0.2} 2 & \cellcolor[rgb]{1,0.8,0.2} 2 & \cellcolor[rgb]{1,0.8,0.2} 1 & \cellcolor[rgb]{1,0.8,0.2} 2 & \cellcolor[rgb]{1,0.8,0.2} 2 & \cellcolor[rgb]{1,0.8,0.2} 2 & \cellcolor[rgb]{1,0.8,0.2} \\
\hline

Monocular RGB Camera & FLIR Firefly S & 1 & 1 & 1 & 1 & 0 & 0 & 1 & 0 & 2 & 1 & \textbf{14} \\

Active 3D (ToF, IR) & Intel D435i & 2 & 2 & 1 & 1 & 1 & 0 & 1 & 2 & 1 & 2 & \textbf{24} \\

Active 3D (ToF, IR) & Intel D605i & 1 & 2 & 1 & 1 & 1 & 0 & 1 & 2 & 0 & 2 & \textbf{20} \\

Active 3D & StereoLabs Zed2 & 2 & 2 & 2 & 2 & 1 & 0 & 1 & 1 & 1 & 1 & \textbf{23} \\

Active 3D  & OAK-D & 1 & 2 & 1 & 1 & 1 & 0 & 1 & 1 & 1 & 1 & \textbf{18} \\
\hline
\end{tabularx}
\end{table*}

%% file: tabfull.tex
\begin{sidewaystable*}
\setlength{\tabcolsep}{6pt}
\setlength{\arrayrulewidth}{0.8pt}
\rowcolors{2}{}{lightgray}
\centering \small
    \caption{Sensors Considered for ReachBot: Full values for LiDAR sensors.}
    \label{tab:full}
\begin{tabularx}{\linewidth}{p{2cm}  p{2cm} *{11}{X} }


\bfseries Sensor Type & 
\bfseries Device Name & 
\bfseries \rotatebox{60}{Resolution} & 
\bfseries \rotatebox{60}{Accuracy} & 
\bfseries \rotatebox{60}{Field of View} & 
\bfseries \rotatebox{60}{Range} & 
\bfseries \rotatebox{60}{Robustness to Darkness} & 
\bfseries \rotatebox{60}{Robustness to Dust} & \bfseries \rotatebox{60}{Power Efficiency} & \bfseries \rotatebox{60}{Implementation Ease} & \bfseries \rotatebox{60}{Mass} & \bfseries \rotatebox{60}{Size} & \bfseries \rotatebox{60}{Affordability} 
\\
\hline
Lidar & RSBPearl & 32 channel 0.2$\degree$ horizontal angular resolution 0.4$\degree$ vertical angular resolution & $\pm\SI{2}{\centi\meter}$ & $360\degree~\times~90\degree$ & \SI{30}{\meter} & High & Low & \SI{13}{\watt} & Low & \SI{920}{\gram} & $\SI{100}{\milli\meter} \times \SI{111}{\milli\meter}$ &  \$3,000-4,000  \\

Lidar & Velodyne Puck (VLP-16) & 16 channel 0.1$\degree$ horizontal angular resolution 0.4$\degree$ vertical angular resolution & $\pm \SI{3}{\centi\meter}$ & $360\degree~\times~30\degree$ & \SI{100}{\meter} & High & Low & \SI{8}{\watt} & High & \SI{830}{\gram} & $\SI{103}{\milli\meter} \times \SI{72}{\milli\meter}$ & \$4,000 \\

Lidar (ToF) & Cygbot Mini Lidar & $160 \times 60$ pixel & $\pm 1\%$ & 65$\degree$ & \SI{0.05}{} - \SI{2}{\meter} & High & Low & - & Low & \SI{28}{\gram} & $\SI{37.4}{\milli\meter} \times \SI{37.4}{\milli\meter} \times \SI{24.5}{\milli\meter}$ & \$170 \\

Lidar & iPhone 12 & - & - & 120$\degree$ & \SI{5}{\meter} & High & Low & - & Low & \SI{164}{\gram} & $\SI{146.7}{\milli\meter} \times \SI{71.5}{\milli\meter} \times \SI{7.4}{\milli\meter}$ & \$999  \\

Lidar & Ouster OS 1-32 & 32 channel 0.35$\degree$  & $\pm \SI{1.5}{\centi\meter}$ & $45\degree~\pm~22.5\degree$ & \SI{120}{\meter} & High & Low & \SI{14}{} – \SI{20}{\watt} & High & \SI{495}{\gram} & $\SI{87}{\milli\meter} \times \SI{74.2}{\milli\meter}$ & \$6,600 \\

Monocular RGB Camera & FLIR Firefly S & $1440\times$ 1080 pixel & - & - & - & Low & Low & \SI{2.2}{\watt} & Low & \SI{20}{\gram} & $\SI{27}{\milli\meter} \times \SI{27}{\milli\meter} \times \SI{14.5}{\milli\meter}$ &  \$234 \\

Active 3D (ToF, IR) & Intel D435i & $1920\times$ 1080 pixel & $<2\%$ at \SI{2}{\meter}  & $87\degree \times 58\degree$ & \SI{0.3}{} - \SI{3}{\meter} & High & Low & \SI{2}{\watt} & High & \SI{260}{\gram} & $\SI{90}{\milli\meter} \times \SI{25}{\milli\meter} \times \SI{25}{\milli\meter}$ & \$334 \\

Active 3D (ToF, IR) & Intel D455i & $1280\times$ 800 pixel & $<2\%$ at \SI{4}{\meter}  & $87\degree \times 58\degree$  &  \SI{0.6}{} - \SI{6}{\meter}  & High & Low & - & High & \SI{380}{\gram} & $\SI{124}{\milli\meter} \times \SI{26}{\milli\meter} \times \SI{29}{\milli\meter}$  & \$419  \\

Active 3D & StereoLabs Zed2 & $1920\times$ 2080 pixel & - & 110$\degree$(H)~$\times$ 70$\degree$(V) $\times$ 120$\degree$(D) & \SI{0.3}{\meter} - \SI{20}{\meter} & High & Low & \SI{1.9}{\watt} & Mid & \SI{166}{\gram} & $175 \times 30 \times 33$ \SI{}{\milli\meter} & \$499 \\

Active 3D  & OAK-D & $1280\times$ 800 pixel & - & $89\degree~\times~80\degree \times 55\degree$ & \SI{0.7}{\meter} - \SI{8}{\meter} & High & Low & - & - & \SI{115}{\gram} & $\SI{110}{\milli\meter} \times \SI{54.5}{\milli\meter} \times \SI{33}{\milli\meter}$ & \$249 \\

Radar & XM132 & - & - & - &  0.5 - \SI{10}{\meter} & - & - & $\SI{0.05}{\milli\watt}$ & Low & \SI{4}{\gram} & $\SI{25}{\milli\meter} \times \SI{20}{\milli\meter}$ & \$20 \\

Radar & TI AWR1843 & 11.3$\degree$ azimuth~$\times 45\degree$ elevation & - & 75$\degree \times 20\degree$ & \SI{12}{\meter} & - & - & - & Low & - & $\SI{10.4}{\milli\meter} \times \SI{10.4}{\milli\meter}$ & \$299 \\
\hline

\end{tabularx}
\end{sidewaystable*}